\documentclass[conf]{new-aiaa}
\usepackage[utf8]{inputenc}

\usepackage{graphicx}
\usepackage{amsmath}
\usepackage[version=4]{mhchem}
\usepackage{siunitx}
\usepackage{longtable,tabularx}
\usepackage{caption}
\usepackage{subcaption}
\usepackage{soul}

\usepackage[numbers]{natbib}
\usepackage{multicol}
\usepackage{pdfpages}
\usepackage{lscape}
\usepackage{wrapfig}
\usepackage{caption}
\newcommand{\etal}{\textit{et al. }}

\title{Using Surprise Index for Competency Assessment in Autonomous Decision-Making}

\author{Akash Ratheesh \footnote{Professional Research Assistant, Ann and H.J. Smead Aerospace Engineering Sciences, University of Colorado Boulder},  
Ofer Dagan \footnote{Graduate Research Assistant, Ann and H.J. Smead Aerospace Engineering Sciences, University of Colorado Boulder}, Nisar R. Ahmed \footnote{Associate Professor, Ann and H.J. Smead Aerospace Engineering Sciences, University of Colorado Boulder}, %Natasha Bosanac \footnote{Assistant Professor, Ann and H.J. Smead Aerospace Engineering Sciences, University of Colorado Boulder}, 
Jay McMahon$^\ddagger$}

\affil{University of Colorado Boulder, 429 UCB, Boulder, CO, 80309}

\begin{document}

\maketitle

\begin{abstract}
This paper considers the problem of evaluating an autonomous system's competency in performing a task, particularly when working in dynamic and uncertain environments. 
The inherent opacity of machine learning models, from the perspective of the user, often described as a `black box', poses a challenge.
To overcome this, we propose using a measure called the Surprise index, which leverages available measurement data to quantify whether the dynamic system performs as expected. 
We show that the surprise index can be computed in closed form for dynamic systems when observed evidence in a probabilistic model if the joint distribution for that evidence follows a multivariate Gaussian marginal distribution.
We then apply it to a nonlinear spacecraft maneuver problem, where actions are chosen by a reinforcement learning agent and show it can indicate how well the trajectory follows the required orbit.
\end{abstract}

% \section{Nomenclature}

% {\renewcommand\arraystretch{1.0}
% \noindent\begin{longtable*}{@{}l @{\quad=\quad} l@{}}
% $E$  & Example \\

% \end{longtable*}}

\section{Introduction}

Over the past few years, there has been increasing use of autonomous systems for various applications, from autonomous driving to aerospace systems. 
These systems are often required to operate in complex and uncertain environments, e.g., the highway or in space. 
For successful deployment and useful operation of these systems around and by humans, there is a need to evaluate the autonomous system's competency in performing its task.  
For this reason \cite{aitken_assured_2014, israelsen_factor-based_2022, conlon_dynamic_2023} present a \emph{Factorized Machine Self-Confidence (FaMSeC)} framework, aimed at allowing a decision-making autonomous system to self-asses its competency.
This paper focuses on one of the factors considered by that framework, the model validity, which is meant to assess how well the underlying model, used to make decisions, reflects reality. 
One key challenge is that the model, often generated by some machine learning (ML) algorithm, might be a ``black box'' with respect to the end-user.
An intuitive solution is to use available data in the form of measurements and assess whether what we see is more surprising than what we expect to see. 
The surprise index (SI) \cite{habbema_models_1976} is a measure that aims to quantify this intuition.
For a set of measurements, it sums up the probability of all other \emph{possible} measurement instances that have a probability less than the observed ones.
A very probable (less surprising) set of measurements will then have more probability mass summed `behind' it.

An important point to note here is that this automatically gives a notion of an equivalence class for the instances of observed measurements based on their probability.
That is, there are `typical sets' of observations or evidence that will not appear surprising to the model, and there will be other `atypical' sets of observations or evidence that will appear to be more surprising.
The (a)typicality of a set of evidence values can be denoted by the probability of seeing available evidence.
Since it is generally difficult to tell apart outliers and extreme events (e.g., for long-tailed distributions) from anomalies that are not explained well by the model, the surprise index allows us to see if a model does not do a good job of explaining the evidence.
% , if we are especially worried about encountering things that should be rare or infrequent more frequently than they ought to be encountered in a particular domain of reasoning (assuming a particular model holds).

This has much in common with NEES/NIS Monte Carlo testing for Kalman filter consistency checking \cite{bar-shalom_linear_2001}. 
That is, we are interested in asking the same question: does the estimated uncertainty of the filter/probabilistic model correctly describe the actual uncertainty encountered in the `real world' implementation for the problem the model is constructed against? 
However, we will show later in this paper that there are subtle differences in how we go about answering that question with the surprise index vs. NEES/NIS tests.

As noted in \cite{zagorecki_approximation_2015, laskey_conflict_1991}, the SI is not easy to compute, and it is generally intractable since it is exponential in the number of evidence variables.
For finitely countable discrete random variable distributions, the SI could be (in principle) evaluated by assigning a total ordering to the outcomes of the random variable, according to the probability of outcome occurrence. 
Then the corresponding cumulative density function (cdf) gives the desired SI. 
In Bayes Nets and other probabilistic models, this may not be so easy to do since the number of observations and possible outcomes of each observation could be prohibitively large. 
Furthermore, sorting by the probabilities requires knowing the probabilities of the joint evidence outcomes in the first place, which
is typically not available in closed form -- indeed, this is what makes Bayesian inference approximations necessary in the first place for such models. To date and to the best of our knowledge, there have not been any results describing how the SI can be exactly computed, even for special cases. The main contributions of this paper are:
\begin{enumerate}
    \item We develop a closed-form solution of SI for a multivariate Gaussian observation model and dynamical system.
    \item We test our closed-form solution on a $2D$ GPS localization example to validate it against the empirical SI  \cite{zagorecki_approximation_2015}.
    \item We demonstrate the application of SI to assess the validity of a new reinforcement learning-based nonlinear spacecraft maneuver planner \cite{bonasera_designing_2023, boone_incorporating_2022}. 
\end{enumerate}

The rest of the paper is organized as follows: in Sec. \ref{sec:bgd}, we motivate the use of SI in the context of ``black box" control systems and give the necessary background on SI.
Sec. \ref{sec:approach} develops a closed-form solution of SI, validates it against an empirical solution, and extends to linear dynamic systems. 
We then apply the solution for a nonlinear spacecraft maneuver scenario in Sec. \ref{sec:results}, and draw conclusions regarding the performance of a reinforcement learning-based control policy. 
Sec. \ref{sec:summary} then presents conclusions and details future research directions.

\section{Background}

Reinforcement learning (RL) based methods are being increasingly explored in the areas of spacecraft control and trajectory planning. 
In real-world spacecraft operations, the true states of the spacecraft are unknown to the user/`trajectory planner,' and maneuvering operations are planned based on an estimate of the spacecraft states. 
Previous works \cite{bonasera_designing_2023, boone_incorporating_2022} have used RL-based techniques to perform maneuver planning for periodic orbits in the Earth-Moon System.
These orbits are unstable and are very sensitive to observation and navigation errors. 
Thus, any errors in the maneuvring action generated by these RL-based planners can have a significant impact on the resulting trajectory. 
Also, the RL-based planners are trained in a model-free manner, and true dynamical models are not available to the planner.
Therefore it is crucial to evaluate these RL agents to validate that the actions generated are `what is expected' from the system. 
Secondly, although a reward function can sufficiently indicate whether a policy is `well-trained,' it may not capture any off-nominal behaviors that can happen during the execution of the trajectory, e.g., deviations due to thruster misfire/malfunction. 

The surprise index can be useful in these scenarios to evaluate these trajectories in an online manner to assess whether the observations of the trajectory resulting from the action match the expected behavior from the system in a probabilistic sense. 
% In order to generate accurate estimates of the spacecraft's states, there exist probabilistic models of the spacecraft's dynamics as well as measurements generated by tracking stations. 
% Surprise Index can leverage these models to evaluate whether the recorded observations/measurements match the observations expected from the system, i.e. whether this sequence of measurements is surprising or not.
Suppose we have some probabilistic model, for instance a Bayes Net (BN) or Markov random field (MRF) or something else which stipulates a joint probability density function (pdf) of the variables of interest, with observed evidence variables $E_{1:N} = \{E_1, E_2,..., E_N\}$.
Let the specific observed instantiation of those evidence variables be denoted $e^{o}_{1:N} = \{e^{o}_1, e^{o}_2,...,e^{o}_N\}$, where $e^o_i \in \text{dom}(E_i)$ and dom is the domain of $E_i$. 
We will use the superscript \emph{o} notation to denote the observed instance and remove the \emph{o} to talk about other hypothetical
instances $e_{1:N} = \{e_1, e_2,...,e_N\}$ which `could' have been observed in principle (this is a key distinction). 
Note that it can be the case that the evidence variables are either discrete, continuous or some mix of both.

The surprise index of an observed instance of evidence variables $e^{o}_{1:N}$ is defined as the sum of the probability of all other events that are less likely to occur,
\begin{equation}
    \mathcal{SI}(e^{o}_{1:N}) = \sum_{e_{1:N}:P(e_{1:N})<P(e^{o}_{1:N})} P(e_{1:N}). 
    \label{eq:SI_def}
\end{equation}
Note that since it sums the probability of all other possible instances $e_{1:N}$ of $E_{1:N}$ which have a probability less than the observed instance $e^o_{1:n}$, it automatically defines the range of the surprise index to be on the interval [0,1], with a higher value indicating less surprise.
In other words, the more probability mass that an observed instance has sitting `behind' it, the less surprising it appears to be, according to the probabilistic model. 

The problem is that, in practice, computing the surprise index requires iterating through all combinations of $e_{1:N}$, which is exponential in the number of evidence variables. Thus it is intractable, as noted in \cite{zagorecki_approximation_2015} and in \cite{laskey_conflict_1991}.
In \cite{zagorecki_approximation_2015}, Zagorecki et al. proposed an asymptotic approximation to the SI, which makes use of the fact that the joint distribution of the evidence follows an approximately log-normal distribution (i.e., the logarithm of the joint evidence probabilities has a Gaussian distribution, owing to the central limit theorem).
Zagorecki et al. claim the approximation is particularly useful when very large $N$ is encountered, i.e., for large networks. 
In \cite{laskey_conflict_1991}, Laskey mentions the SI in the context of original work by Habbema \cite{habbema_models_1976}, but essentially goes no further than saying that it is intractable to compute, proceeding instead to look at different heuristics such as Jensen \etal's conflict score \cite{jensen_analysis_1990}, which is closely related to the mutual information of the evidence variables.

Nevertheless, as discussed in Zagorecki \etal \cite{zagorecki_approximation_2015}, the surprise index has several useful properties that make it a useful measure for model quality assessment in machine competency self-assessment. These properties also make it preferable in many respects to alternative goodness of fit type measures for probabilistic models, such as log-likelihood scores (which can become unboundedly large for a large number of observations) or entropy (for which minimum and maximum values depend on the number of outcomes described by a probability distribution). Unlike these measures, the surprise index retains an interpretable meaning as a probability and so always takes on the range of 0 to 1 regardless of the number of outcomes. Moreover, the surprise index has a very close connection to the concept of p-values from statistics, which are also widely used for goodness of fit testing. Unlike p-values, however, the surprise index does not require hypothesis tests to attach a specific meaning to its calculation. 

In the next section, we show that the surprise index can actually be computed in closed form for observed evidence in a probabilistic model if the joint distribution for that evidence follows a multivariate Gaussian marginal distribution. 
The significance of this is to connect the surprise index to the well-known chi-square hypothesis tests for things like Kalman filters (NIS tests for filter consistency, specifically) \cite{bar-shalom_linear_2001}. 
As we'll explore later, the surprise index may provide a useful complementary way of assessing `goodness of fit' (g.o.f.) for a probabilistic model that does not rely on the ability of that model to actually process data (i.e., do inference, which the NIS test for Kalman filters does require). 
Hence, it could be a useful alternative metric for tuning things like Kalman filters, extended Kalman filters, and other probabilistic inference tools for estimation and control (e.g., LQG, RL).

% \subsection{\hl{Halo-Orbit Scenarios \& RL policy}}
% \begin{itemize}
%     \item Refer the RL approach in the Spencer's Paper
%     \item Halo Orbit \& Scenarios - Station Keeping \& Reconfiguration
%     \item Why SI is useful 

% \end{itemize}
\label{sec:bgd}

\section{Approach and Methodology}
\label{sec:approach}

\begin{figure}[tb]
    \centering
    \includegraphics[width=0.45\linewidth]{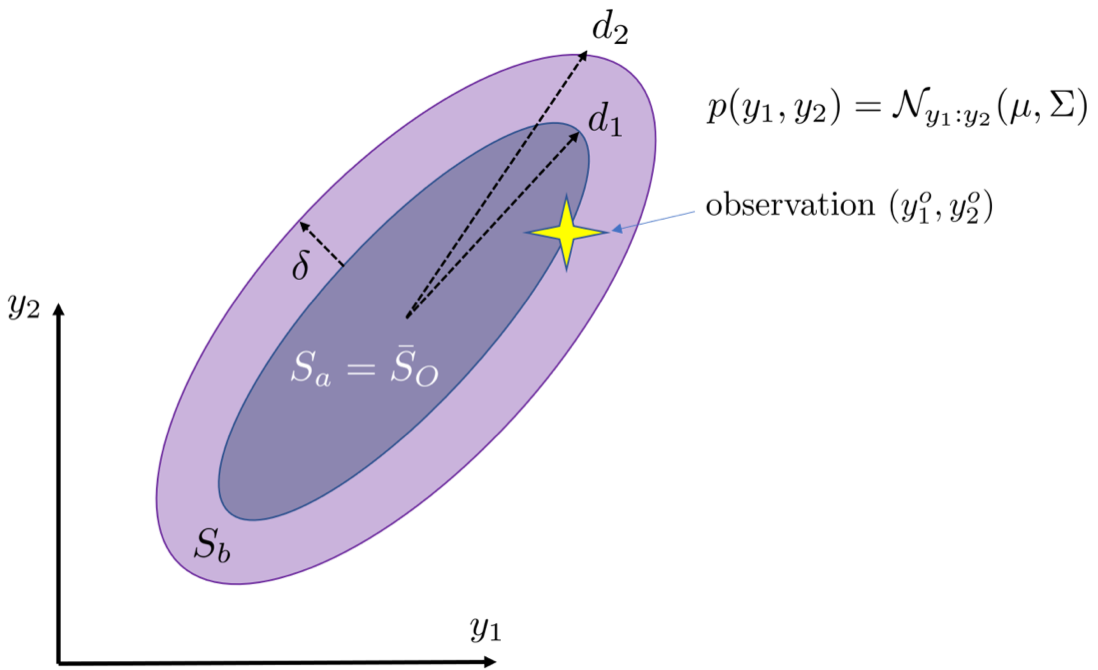}
    \caption{Ordered probabilistic equivalence sets for Gaussian pdf with $N=2$.}
    \label{fig:levelSets}
    % \vspace{-0.2in}
\end{figure}

Consider a probabilistic model where the evidence/observation distribution is a continuous N-variate Gaussian pdf. 
Suppose we observe random vectors $E_i = Y_i \in \mathbb{R}^{p \times 1}$, so that $E_{1:N}  = Y_{1:N}$ can be represented as a $p \times N$ array of vectors, where 
\begin{equation}
    Y_{1:N} \sim \mathcal{N}_{y_{1:N}}(\mu,\Sigma).
    \label{eq:generalpdf}
\end{equation}
We can consider the realization, $y_{1:N}$ to be a concatenated vector of $p \times 1$ random vector stacked in to $N$ blocks, where $\mu \in \mathbb{R}^{Np \times 1}$ and $\Sigma \in \mathbb{R}^{Np \times Np}$. 
This describes the joint distribution for the possible sets of $y_{1:N}$ instances of $Y_{1:N}$.  
Similarly, the pdf for the realized (observed) instances $y^o_{1:N}$ would be

\begin{equation}
    P(y^o_{1:N}) = \mathcal{N}_{y^o_{1:N}}(\mu,\Sigma).
    \label{eq:evidencepdf}
\end{equation}
% .... (Should include hypercube?)
In order to calculate SI according to Eq. (\ref{eq:SI_def}), we need to order all possible sets of observations $E_{1:N}\in S$ and sum/integrate the ones that have less probability than the observed set $E^o_{1:N}\in S^o$.
For example, if we consider two non-overlapping regions for integration $S_a$ and $S_b$, where the realizations for $Y_{1:N}$ in $S_a$ are equally more probable (less surprising) than the realizations in $S_b$ (more surprising), then
\begin{equation} 
    p(y_{1:N} \in S_a) = \int_{S_a} \mathcal{N}_{y_{1:N}}(\mu,\Sigma) dy_{1:N} > p(y_{1:N} \in S_b) = \int_{S_b} \mathcal{N}_{y_{1:N}}(\mu,\Sigma) dy_{1:N}.
    \label{eq:region_integration}
\end{equation}
Fortunately, for multivariate Gaussian pdf, such ordered regions do, in fact, exist, and their boundaries are given by the level sets defined by the Mahalanobis distance. 
The  Mahalanobis distance $d_M$ for the joint realization vector $y_{1:N}$ is given by
\begin{equation}
    d_M = \frac{1}{2}(y_{1:N} - \mu)^T\Sigma^{-1}(y_{1:N} - \mu).
    \label{eq:mahalDist}
\end{equation}
To demonstrate, suppose that $d_1 > 0$ and $d_2 = d_1 + \delta$ for some small $\delta \in [0,d_1]$, we can define $S_a$ and $S_b$ as follows
\begin{equation}
    \begin{split}
        &S_a = \{y_{1:N}: d_M \le d_1\} \\
        &S_b = \{y_{1:N}: d_1 \le d_M \le d_2\},
    \end{split}
    \label{eq: levelSetDef}
\end{equation}
then for $N=2$ computing the probabilities according to Eq. (\ref{eq:region_integration}) leads directly to the dark and lighter purple regions shown in Fig. \ref{fig:levelSets}, corresponding to $S_a$ and $S_b$, respectively. 
Note that this readily extends to higher dimensions, where $S_a$ and $S_b$ are defined as the contours of hyper-ellipses in $\mathbb{R}^{N_p\times 1}$ about the mean $\mu$, with orientation defined by $\Sigma$.

The question now is how to compute the probabilities and resulting SI for a specific set of observations $y_{1:N}^o$ that have a joint Gaussian pdf.
We know that the quadratic form for Gaussian random variables given by the Mahalanobis distance results in a chi-square distributed scalar random variable with $Np$ degree of freedom \cite{bar-shalom_linear_2001},
\begin{equation}
    \epsilon = (y_{1:N} - \mu)^T\Sigma^{-1}(y_{1:N} - \mu) \sim \chi^2_{\epsilon}(Np).
    \label{eq:chiSquare}
\end{equation}
Notice that the chi-square variable $\epsilon$ and the Mahalanobis distance $d_M$ (Eq. (\ref{eq:mahalDist})) are related through $d_M=\frac{1}{2}\epsilon$, or $\epsilon=2d_M$.
Thus, all possible observation sets, $S_r$, that are less probable than the observed set $y_{1:N}^o$, will have larger Mahalanobis distance, i.e. larger $\epsilon$. 
But since $S_r$ is unbounded, we can instead look at the complimentry\footnote{This evokes the commonly used `observation gating probability' found in the data association literature, e.g., for probabilistic data association methods in multi-target tracking \cite{bar-shalom_probabilistic_2009}}, more probable, set $\bar{S}_O = \{ y_{1:N}: d_M(y_{1:N}) \le d_M(y^o_{1:N}) \}=\mathbb{R}^{Np \times 1} - \bar{S_r}$.
In Fig. \ref{fig:levelSets}, if $d_M(y^o_{1:N})=d_O=d_1$, this means that instead of considering all level sets that lie outside of $d_1$, i.e., for different $S_b$, we consider the ones inside $S_a=\bar{S}_O$.
It follows that

\begin{equation}
    \begin{split}
        \mathcal{SI}(y^o_{1:N}) &= P(y_{1:N} \in S_r) = \int_{S_r} \mathcal{N}_{y_{1:N}}(\mu,\Sigma) dy_1...dy_N
        = 1-P(y_{1:N} \in S_O)\\ &= 1 - \int_{S_O} \mathcal{N}_{y_{1:N}}(\mu,\Sigma) dy_1...dy_N
        = 1-P(\epsilon<\epsilon_O)=1-\int^{\epsilon_O}_0\chi^2_{\epsilon}(Np)d\epsilon.
        \label{eq:SI}
    \end{split}
\end{equation}

Note that this is just the complementary cdf of the chi-square distribution. 
Also note that the maximum value of $\mathcal{SI}=1$ (least surprising) is achieved at $y^o= \mu$, whereas the minimum value of $\mathcal{SI}=0$ (most surprising) is achieved asymptotically as the elements of $y^o$ approach $\pm\infty$.

At this point, it is worth re-stating the interpretation of the surprise index: it denotes the probability of obtaining any realization for $Y_{1:N}$ that is less probable than the observed realization sequence $y^o_{1:N}$ under the assumed model. 
Hence, for a jointly Gaussian distributed set of observations, the probability of obtaining any other possible observation realization that is less probable than a given observation decreases outwardly from the mean, in agreement with intuition.

\subsection{Validation Example - GPS Localization}

Suppose we have a 2-D GPS-like localization sensor that provides noisy measurements $y_k \in \mathbb{R}^2$ for a stationary platform located at position $x \in \mathbb{R}^2$. Consider that the measurements generated by the sensor are defined by the following prior probabilistic model
\begin{equation}
    y_k = x + v_k, \quad x \sim \mathcal{N}(\mu_{0},\Sigma_0), \quad v_k \sim \mathcal{N}(0,R).
\end{equation}
Here, $\mu_{0}$ is the prior mean of the platform's position, $\Sigma_0$ is the prior covariance in the platform position, and $R$ is the true discrete-time additive white Gaussian noise (AWGN) covariance of the sensor.

\begin{figure}[b]
    \centering
    \begin{subfigure}[b]{0.45\textwidth}
         \centering
         \includegraphics[width=\textwidth]{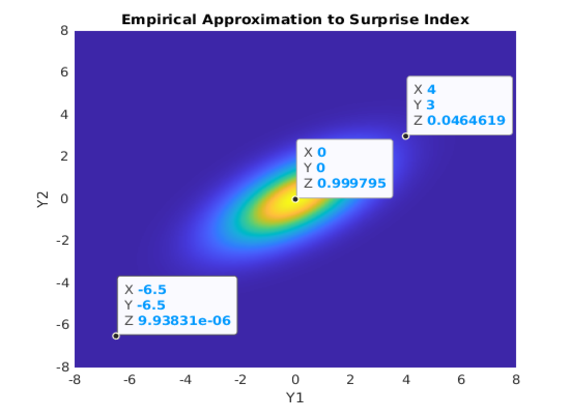}
         \label{fig:y equals x}
     \end{subfigure}
     \hfill
     \begin{subfigure}[b]{0.45\textwidth}
         \centering
         \includegraphics[width=\textwidth]{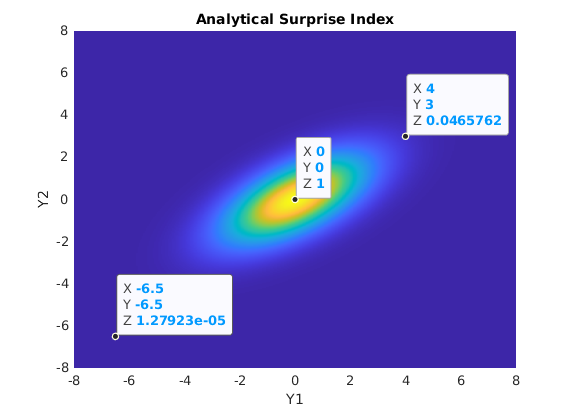}
         \label{fig:three sin x}
     \end{subfigure}
    \caption{Empirical grid approximation of SI according to \cite{zagorecki_approximation_2015} (left) and the new analytical SI (right).}
    \label{fig:gps_example1}
    \vspace{-0.2in}
\end{figure}

Now suppose we record a fixed-length realization of $Y_{1:N}$, $y^o_{1:N} = \{y^o_1, ... ,y^o_N\}$. In order to calculate the Surprise Index for the successive sequence $y^o_{1:k}$ for $1 \le k \le N$, we have to obtain the prior probability of observing sequences  $y_{1:k}$ that are less probable than $y^o_{1:k}$. Figure \ref{fig:gps_example1} shows the comparison between the SI calculated using a grid-based brute force approximation -- not dissimilar to the empirical method described in \cite{zagorecki_approximation_2015} -- and the analytical method we developed (\ref{eq:SI}). We can see good agreement between the two methods, thus giving initial validation to our new analytical solution.

Note that by the definition of the SI, we do not need to filter or process the data at all to produce a state estimate for $x$. 
Rather, we `just' need to assess the a priori probability of obtaining sequences $y_{1:k}$ that are less probable than $y^o_{1:k}$. 
This is important to consider in the context of assessing models for processing data with techniques like batch least-squares or Kalman filtering since typical `residual' statistics for these methods are based on measures of a posteriori fit or surprise (NEES/NIS). 
% \hl{We will extend on the subtle, yet important, differences between SI and NEES/NIS tests in the final manuscript.}

\subsection{Generalization to Dynamic Filtering Problems} 
\label{section:dynamic}
We derive the general expressions for computing the SI with a discrete-time linear Gaussian state space system, i.e., a linear Gauss-Markov process.
Consider a discrete-time linear Gaussian state space system with known single-step state transition matrix $F_k$, control parameter $G_k$, process injection parameter $\Gamma_k$ and observation matrix $H_k$,
\begin{equation}
    \begin{split}
        &x_{k+1} = F_k x_k + G_k u_k + \Gamma_k w_k, \quad w_k \sim \mathcal{N}(0,Q_k)\\
        &y_{k+1} = H_{k+1}x_{k+1} + v_{k+1}, \quad v_{k+1} \sim \mathcal{N}(0,R_{k+1}),
    \end{split}
    \label{eq:dynamics}
\end{equation}
where $x$ is the state, $y$ is the sensor measurements, $u_k$ are the known control inputs at time step $k$, $w_k$ and $v_{k+1}$ are AWGN process noise and measurement noise with covariance matrix $Q_k$ and $R_{k+1}$ respectively.
Suppose the initial states of the system are unknown and modeled as Gaussian prior $x_0 = \mathcal{N}(\mu_{x_0},\Sigma_{x_0})$ where $\mu_{x_0}$ is the mean and $\Sigma_{x_0}$ is the apriori covariance.

We need to compute the marginal joint pdf of the observation sequence vector $\bar{y}_{1:N} \equiv [\bar{y}^T_1, ...,\bar{y}^T_N]$,
\begin{equation}
    p(\bar{y}_1, ... ,\bar{y}_N) = \mathcal{N}_{y_{1:N}}\left( \mu_{y_{1:N}}, \Sigma_{y_{1:N}} \right).
\end{equation}
If we assume that the system does not have any process noise, we can compute the expected measurement $\mu_{y_k} \in  \mu_{y_{1:N}}$ for $1 \le k \le N$ by,
\begin{equation}\label{eq:mu_1_n}
    \mu_{y_k} = H_k x_k = H_k [ \Phi(0,k)\mu_{x_0} ]
\end{equation}
where $\Phi$ is the state transition function,
\begin{equation}
    \Phi(k-T,k) = \prod^T_{i=1}F_{k-i}, \quad \Phi(k,k) = I
\end{equation}

In order to compute the marginal joint pdf of the observation sequence, we also need to find the covariance matrix $\Sigma_{y_{1:N}}$. The block diagonal and block off-diagonal term of $\Sigma_{y_{1:N}}$ is given by
\begin{equation}\label{eq:Sigma_1_n_diagonal}
\begin{split}
    \mathbb{E}\left[(y_i - \mu_i)(y_i - \mu_i)^T\right]  = \mathbb{E}[y_i y_i^T] - \mu_{yi}\mu^T_{yi} 
     = R_i + H_i\left[\Phi(0,i) \Sigma_{x_0} \Phi^T(0,i)\right]H^T_i,
\end{split}
\end{equation}
\begin{equation}\label{eq:Sigma_1_n_off_diagonal}
\begin{split}
    \mathbb{E}\left[(y_i - \mu_i)(y_j - \mu_j)^T\right]  = \mathbb{E}[y_i y_j^T] - \mu_{yi}\mu^T_{yj} 
     = H_i\left[\Phi(0,i) \Sigma_{x_0} \Phi^T(0,j)\right]H^T_j.
\end{split}
\end{equation}
Now, if the system is modeled with process noise, Eqs. \ref{eq:mu_1_n},\ref{eq:Sigma_1_n_diagonal}, \ref{eq:Sigma_1_n_off_diagonal} becomes

\begin{equation}
    \mu_{y_k} = H_k \left[ \Phi(0,k)\mu_{x_0} + \sum_{i=1}^k \Phi(k+1-i,k)G_{k-1}u_{k-1}\right],
\end{equation}

\begin{equation}
    \mathbb{E}\left[(y_i - \mu_i)(y_i - \mu_i)^T\right] = R_i + H_i\left[[\Phi(0,i) \Sigma_{x_0} \Phi^T(0,i) + \sum_{d=0}^{i-1} \Phi(d+1,i)\Gamma_d Q_d \Gamma_d^T \Phi^T(d+1,i) \right]H_i^T,
    \label{eq:Py_diag}
\end{equation}

\begin{equation}
    \mathbb{E}\left[(y_i - \mu_i)(y_j - \mu_j)^T\right] = H_i\left[\Phi(0,i) \Sigma_{x_0} \Phi^T(0,j) + \sum_{d=0}^{m=min(i-1,j-1)} \Phi(d+1,i)\Gamma_d Q_d \Gamma_d^T \Phi^T(d+1,j)\right]H^T_j.
    \label{eq:Py_offdiag}
\end{equation}
With this derivation, we are able to analytically compute the SI for dynamic systems and apply it to validate complex models.

\section{Example Problem - Spacecraft Maneuver}
\label{sec:results}

A surprise index can be useful for detecting off-nominal behavior from the system. 
While previous works on SI rely on empirical/sampling-based methods to approximate the SI, which tends to be computationally intractable, we have shown here that when the measurement model follows a multivariate Gaussian probabilistic model, we can find a closed-form solution for SI.  

We now show how SI can be used to analyze the decision-making or policy of an RL-based model.
More specifically, we test it for spacecraft stationkeeping scenarios around halo orbits in the Sun-Earth system.
% maneuver planner applications, which operate in unstable orbit scenarios such as station keeping around L2 Halo orbits and L2 Lyapanov-to-Halo Orbit transfers.  
In these scenarios, the RL policy is trained to correct any deviation caused by perturbations and other uncertainties, e.g., observation and navigation errors \cite{boone_incorporating_2022}. 
The RL policy is trained in the circular restricted three-body problem (CR3BP); for details on the dynamical model, training process, and required parameters, see \cite{boone_incorporating_2022}.
We simulate spacecraft trajectories, operating under policy $\pi_2$, where the input to the policy is a noisy measurement of the state, and the sensor uncertainty is accounted for in the training process.
The trajectories are perturbed around the working point, where the agent was trained vs. further from this working point, to test the robustness of the policy.
In reality, these perturbations may occur due to imperfect models/data used to train the RL policy or due to other errors and uncertainties that might occur during the actual execution of the maneuvers.

SI can be used to evaluate these policies by leveraging the generalized dynamic formulation described in Section \ref{section:dynamic}.
As in \cite{boone_incorporating_2022}, we assume that the system has no process noise and that we have a sensor with full observability of the spacecraft.
The (linear) measurement model is defined according to eq. (\ref{eq:dynamics}), with measurement matrix $H=\mathcal{I}_{6\times 6}$.
With these assumptions, equations (\ref{eq:Py_diag})-(\ref{eq:Py_offdiag}) become,
\begin{equation}
    \mathbb{E}\left[(y_i - \mu_i)(y_i - \mu_i)^T\right] = \Sigma_{y_{i}} = R_i + \Phi(0,i) \Sigma_{x_0} \Phi^T(0,i)=R_i+\Sigma_{x_i},
    \label{eq:Py_diag_maneuver}
\end{equation}

\begin{equation}
    \mathbb{E}\left[(y_i - \mu_i)(y_j - \mu_j)^T\right] = \Sigma_{y_{i},y_j} =\Phi(0,i) \Sigma_{x_0} \Phi^T(0,j)=\Sigma_{x_{i},x_j}.
    \label{eq:Py_offdiag_maneuver}
\end{equation}

The challenge, in this case, is that the system's dynamics are highly nonlinear, thus, a closed-form expression for the state transition matrix $\Phi(0, i)$ does not exist. 
Instead, the dynamics of the system are defined by the nonlinear function $f(x,u)$, and while an approximation for $\Phi(0,i)$ can be calculated through the Jacobian of $f(x,u)$, it provides a poor approximation for uncertainty propagation. 
Another method to propagate uncertainty through nonlinear systems is by the sigma point or unscented transformation \cite{julier_new_1995}.

\subsection{Uncertainty Propagation via Sigma Points}
\label{subsec:SP}
The sigma point transformation allows for approximating the first two moments (mean and covariance) of the distribution by propagating $2N+1$ `sigma points' through the nonlinear dynamics, where $N$ is the number of states.
After propagating the points, the new mean and covariance can be calculated using,
\begin{equation}
    \begin{split}
        \mu_x &= \sum_{n=0}^{2N}\alpha_n\chi_n \\
        \Sigma_{x} & = \sum_{n=0}^{2N}\alpha_n(\chi_n-\mu_x)(\chi_n-\mu_x)^T,  
    \end{split}
    \label{eq:sigPoints}
\end{equation}
where $\chi_n$ is the $n$-th sigma point, and $\alpha_n$ is a weighting factor, such that $\sum \alpha_n=1$. For more details on calculating the sigma points and the weighting factor, see \cite{julier_new_1995}.
While the above equation can be used to calculate the covariance in eq. (\ref{eq:Py_diag}), this accounts for only the main diagonal blocks of the joint pdf. 
To calculate the off-diagonal blocks or correlations between measurements taken in different time steps, $\Sigma_{y_{i},y_j}$, according to eq. (\ref{eq:Py_offdiag}), we compute the following, 
\begin{equation}
        \Sigma_{x_{i},x_j} = \sum_{n=0}^{2N}\alpha_n(\chi_{n_i}-\mu_{x_i})(\chi_{n_j}-\mu_{x_j})^T.
    \label{eq:sigPoints_cross}
\end{equation}
Here $\chi_{n_i}$ ($\chi_{n_j}$) and  $\mu_{i}$ ($\mu_{j}$) are the sigma points and means at time step $i$ ($j$), respectively.

\subsection{Test Case 1 - Nominal Conditions}
\begin{figure}
    \centering
    \includegraphics[width=0.9\linewidth]{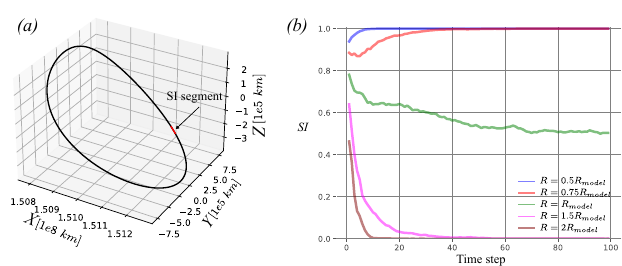}
    \caption{(a) Reference orbit, with red indicating the segment of the orbit that the SI was calculated over. (b) SI results for different true sensor uncertainty values relative to the assumed sensor model.}
    \label{fig:nom_scenario}
\end{figure}

To verify that the SI calculation method makes sense in the case of a nonlinear dynamic system, we first apply it to a nominal case.
In this scenario, the spacecraft exactly follows the reference trajectory depicted in Fig. \ref{fig:nom_scenario}.

% \begin{figure}[tb]
%     \centering
%     \begin{subfigure}[b]{0.45\textwidth}
%          \centering
%          \includegraphics[width=\textwidth]{figures/nom_orbit.pdf}
%          % \label{fig:nom_ref_traj}
%      \end{subfigure}
%      \hfill
%      \begin{subfigure}[b]{0.45\textwidth}
%          \centering
%          \includegraphics[width=\textwidth]{figures/cummulativeSI_noAction.pdf}
%          % \label{fig:nom_SI_results}
%      \end{subfigure}
%     \caption{Left: Reference orbit, with red indicating the segment of the orbit that the SI was calculated over. Right: SI results for different true sensor uncertainty values relative to the assumed sensor model.}
%     \label{fig:nom_scenario}
%     % \vspace{-0.2in}
% \end{figure}
We then perform 5 different 100 Monte-Carlo (MC) simulations, where we change the sensor model statistics $R$ for each batch of simulations.
The presented SI is the cumulative SI, based on all measurements from time 0 to time step $k$.
The results demonstrate the expected, intuitive SI for the different models: when the `true' sensor is better, i.e., has a lower $R$ value, than the assumed sensor model, as in the case of $R=0.5R_{model}$ and $R=0.75R_{model}$, SI quickly converges to 1, meaning that the results are not surprising.
On the other hand, when the true sensor is worse than the assumed model, as in the case of $R=1.5R_{model}$ and $R=2R_{model}$, the system sees surprising measurements, and SI goes to zero. 
The most interesting result is when the true sensor and the model agree, $R=R_{model}$, as shown by the green line.
In this case, the distribution for the SI value at any given time step should be uniform on the range $[0,1]$. Thus, the mean should be $0.5$, as achieved from the 100 MC simulations.

\subsection{Test Case 2 - RL Policy Evaluation}
In this test case, we evaluate the SI using the RL policy $\pi_2$ developed in \cite{boone_incorporating_2022}. 
We perform 3 sets of 100 MC simulations, using sensor statistics used for training and sampling an initial state in simulation.
In each simulation, the initial state is sampled from a normal distribution with a mean equal to the nominal trajectory and a different covariance for each set of simulations.
After the initial state is sampled, a noisy measurement $y_0$ is taken according to eq. (\ref{eq:dynamics}).
The noisy measurement is given as input to the RL policy $\pi_2$, which provides an action (a change in velocity).
This process of taking a noisy measurement and using it as input to the policy $\pi_2$, repeats 15 times along one time period of the halo orbit. 

\begin{figure}[tb]
    \centering
    \includegraphics[width=0.9\linewidth]{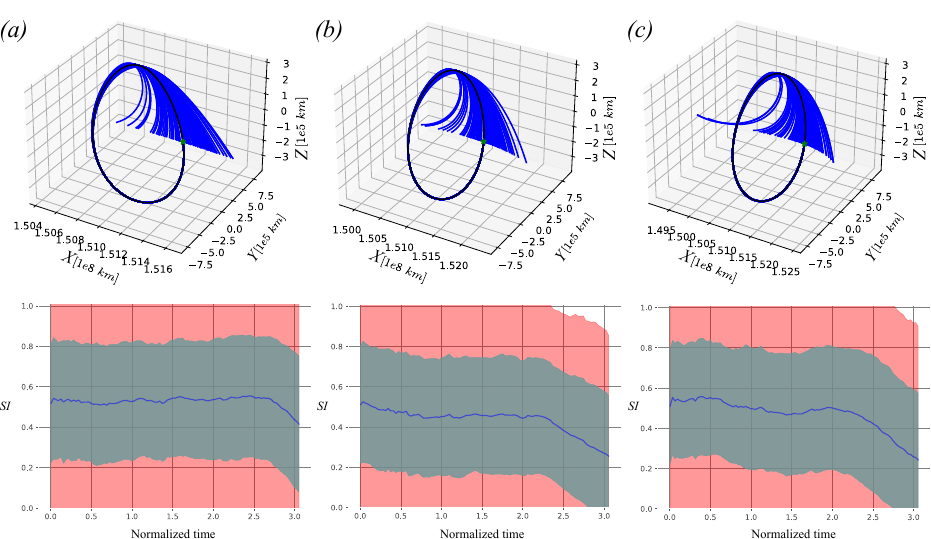}
    \caption{Trajectories (upper row) and SI evaluation (lower row) for stationkeeping scenario with different initial uncertainty. (a) Initial uncertainty is equivalent to what the policy was trained on $P_0$. (b) and (c): $3$ and $5$ times the trained initial uncertainty, respectively.   }
    \label{fig:policyEval}
    \vspace{-0.2in}
\end{figure}

Results of the MC simulations are given in Fig. \ref{fig:policyEval}.
Where Fig. \ref{fig:policyEval}(a) uses a covariance equal to the trained initial covariance, $P_0=P_{trained}$.
Figures \ref{fig:policyEval}(b)-(c) use a larger initial covariance of $3P_0$ and $5P_0$, respectively, i.e., sampling a more deviated initial state, outside of distribution of the trained policy.
The upper row of Fig. \ref{fig:policyEval} shows the reference (nominal) orbit in black and 100 MC trajectories in blue. 
We can see that in all cases, the trajectory follows the reference orbit for more than $2/3$ of the period.
The SI in the lower row of the plots shows the mean (in blue) and $1\sigma$ and $2\sigma$ regions of the 100 simulations in shaded gray and pink, respectively.
Since evaluating the SI at every time step has a significant computational burden due to inversion of the joint covariance matrix (eq. (\ref{eq:chiSquare}), the SI was evaluated using 100 measurements equally spaced along the trajectory.
The SI results show that when the trajectory aligns with the expected orbit, the mean is about 0.5, as we saw in the nominal case (green line in Fig. \ref{fig:nom_scenario}).
However, when the trajectories start to diverge, the SI value drops, which happens earlier when the initial uncertainty is higher. 

\begin{figure}[tb]
    \centering
    \includegraphics[width=0.9\linewidth]{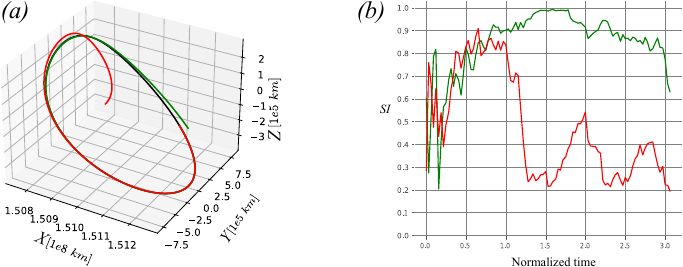}
    \caption{Trajectory comparison -- comparing good performance (green trajectory) to a deviated trajectory (red). (a) shows the trajectories vs. the reference (black) stationkeeping orbit. (b) shows the corresponding SI. }
    \label{fig:traj_comp}
    \vspace{-0.2in}
\end{figure}

Figure \ref{fig:traj_comp} shows two of the 100 trajectories in Fig. \ref{fig:policyEval}(a), i.e. with initial uncertainty $P_0$.
We can see that the SI value indicates deviation from the reference trajectory, where the green trajectory follows the orbit most of the time, with $SI>0.8$, and when it starts diverging at the end, the SI plummets. 
The red trajectory starts diverging much earlier, which is also shown by the values of the SI.

% \textbf{Scenarios to explore:}
% \begin{itemize}
%     \item \st{Perfect alignment of model and truth: show SI=1 for cumulative SI, and random for only current SI. Show the correlation to the actual measurement sigma on the other axis.} 
%     \item \st{Sensor model is better than the actual model (the trajectory is ok): show that SI$\leq 1$.}
%     \item Policy evaluation: 
%     \begin{enumerate}
%         \item benchmark with statistics as the ones that the model was trained with.
%         % Starting from a different initial condition, but the same control as the nominal.
%         \item Sample initial state from $\Sigma_0$ larger than the one that was trained.
%         % Starting from a different initial condition, with noisy measurement and use the policy.
%         \item Same initial uncertainty as the trained one, but different sensor model ($R$).
%     \end{enumerate}
% \end{itemize}

\section{Summary}
\label{sec:summary}
As autonomous systems' decision-making relies more on `black-box' policies generated by a learning algorithm, the surprise index provides a possible avenue to assess the system's competence. 
That is, assess whether the performance of the autonomous system agrees with what we expect it to do.
In this paper, we derived a closed-form solution to calculate the SI in the case of multivariate Gaussian measurement models. 
We applied it to the nonlinear system of the spacecraft stationkeeping maneuver and showed that it provides a good measure of the spacecraft's performance against a reference model of its expected behavior. 

However, computing the SI value for large systems or over a long time horizon might be computationally intractable since it requires inverting the joint covariance matrix over all measurements.
Thus future work will explore approximation and/or efficient calculation methods for SI.

% \section{Conclusion}
% A conclusion section is not required, though it is preferred. Although a conclusion may review the main points of the paper, do not replicate the abstract as the conclusion. A conclusion might elaborate on the importance of the work or suggest applications and extensions. \textit{Note that the conclusion section is the last section of the paper that should be numbered. The appendix (if present), acknowledgment, and references should be listed without numbers.}

% \section*{Appendix}

% An Appendix, if needed, should appear before the acknowledgments.
% \newpage
\section*{Acknowledgments}
The authors acknowledge the support of Prof. Natasha Bosanac of CU Boulder. This work is supported by an Early Stage Innovations grant from NASA’s Space Technology Research Grants Program, under NASA grant 80NSSC19K0222.

\bibliography{references}

\end{document}